\documentclass[10pt,journal,compsoc]{IEEEtran}
\usepackage{graphicx}
\usepackage{multirow}
\usepackage{bm}
\usepackage{float}
\usepackage{pdfpages}
\usepackage{pgfplots}
\usepackage{subfigure}
\usepackage{tikz}
\usepackage{comment}
\usepackage{amsmath,amssymb} 
\usepackage{color}

\ifCLASSOPTIONcompsoc
  \usepackage[nocompress]{cite}
\else
  \usepackage{cite}
\fi
\ifCLASSINFOpdf
\else
\fi
\begin{document}
%
\title{Gradient-based Point Cloud Denoising with Uniformity}
%
%
%
%

\author{Tian-Xing Xu,
        Yuan-Chen Guo,
        Yong-Liang Yang, 
        and Song-Hai Zhang 
\IEEEcompsocitemizethanks{\IEEEcompsocthanksitem Tian-Xing Xu, Yuan-Chen Guo, and Song-Hai Zhang are with the Department
of Computer Science and Technology, Tsinghua University, Beijing, China, 100084. 
\IEEEcompsocthanksitem Yong-Liang Yang is with Department of Computer Science, University of Bath, Bath, United Kingdom, BA2 7AY.
\IEEEcompsocthanksitem This work has been submitted to the IEEE for possible publication. Copyright may be transferred without notice, after which this version may no longer be accessible.
}

}

%
%

\markboth{Journal of \LaTeX\ Class Files,~Vol.~14, No.~8, August~2015}%
{Shell \MakeLowercase{\textit{et al.}}: Bare Demo of IEEEtran.cls for Computer Society Journals}
%



\IEEEtitleabstractindextext{%
\begin{abstract}
Point clouds captured by depth sensors are often contaminated by noises, obstructing further analysis and applications. In this paper, we emphasize the importance of point distribution uniformity to downstream tasks. We demonstrate that point clouds produced by existing gradient-based denoisers lack uniformity despite having achieved promising quantitative results. To this end, we propose GPCD++, a gradient-based denoiser with an ultra-lightweight network named UniNet to address uniformity. Compared with previous state-of-the-art methods, our approach not only generates competitive or even better denoising results, but also significantly improves uniformity which largely benefits applications such as surface reconstruction. 
\end{abstract}

\begin{IEEEkeywords}
Point cloud, denoising, gradient fields, uniformity.
\end{IEEEkeywords}}

\maketitle

\IEEEdisplaynontitleabstractindextext

%
\IEEEpeerreviewmaketitle

\IEEEraisesectionheading{\section{Introduction}\label{sec:introduction}}

%
%
%
%
\IEEEPARstart{W}{ith} the rapid development of 3D data acquisition devices such as depth cameras and LiDAR sensors, 3D point clouds have drawn increasing attention. Many applications that rely on such irregular data have grown in recent years, including point-cloud-based semantic understanding~\cite{ReDAL,RPVNet,NoisyLabelSeg,PerturbedSelfDistillation,TempNet}, 3D surface reconstruction~\cite{surface_rec_2,PolyFit,surface_rec_1}, and point cloud rendering~\cite{point_cloud_rendering,guennebaud2007algebraic}. However, raw point clouds acquired directly from sensors are often corrupted by noises due to hardware constraints, which largely affects downstream tasks. Therefore, point cloud denoising is often employed as a post-processing step to remove the noises and generate "clean" point clouds.

  

\begin{figure*}[!t]
    \centering
    \hspace{-0.5cm}
    \subfigure[Noisy Input]{
        \begin{minipage}[t]{0.19\linewidth}
            \centering
            \includegraphics[width=\linewidth]{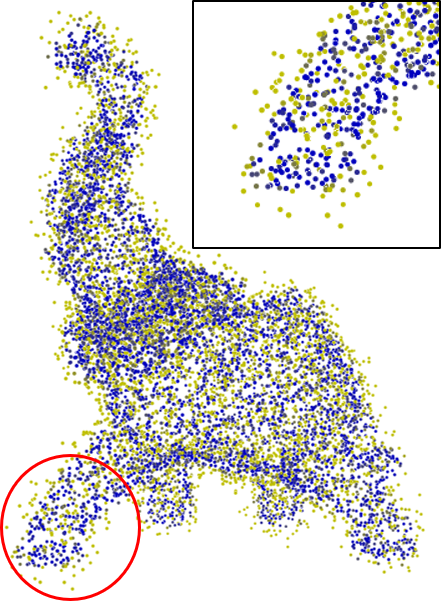}
        \end{minipage}
    }
    \hspace{-0.22cm}
    \subfigure[Ground Truth]{
        \begin{minipage}[t]{0.19\linewidth}
            \centering
            \includegraphics[width=\linewidth]{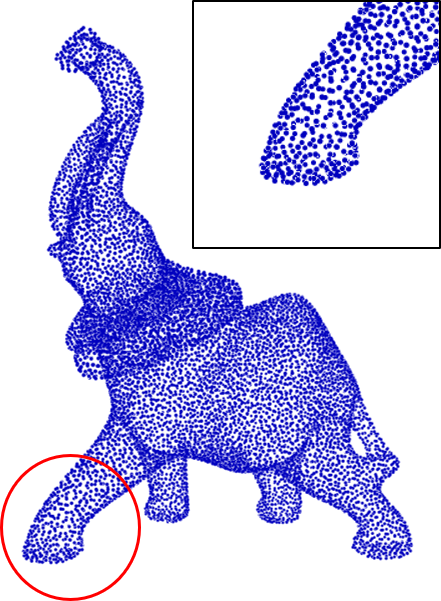}
        \end{minipage}
    }
    \hspace{-0.22cm}
    \subfigure[Score-based~\cite{luo2021score}]{
        \begin{minipage}[t]{0.19\linewidth}
            \centering
            \includegraphics[width=\linewidth]{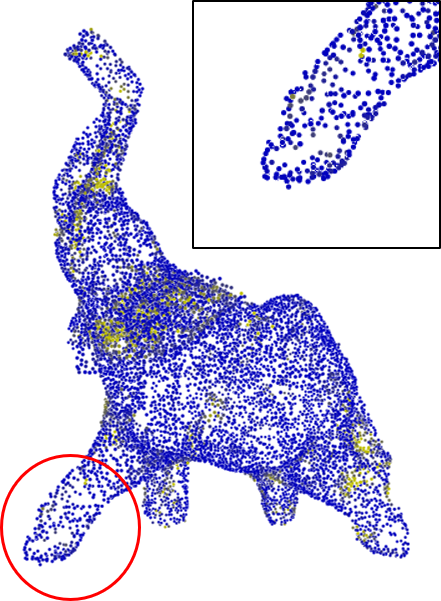}
        \end{minipage}
    }
    \hspace{-0.22cm}
    \subfigure[PSR~\cite{chen2021deep}]{
        \begin{minipage}[t]{0.19\linewidth}
            \centering
            \includegraphics[width=\linewidth]{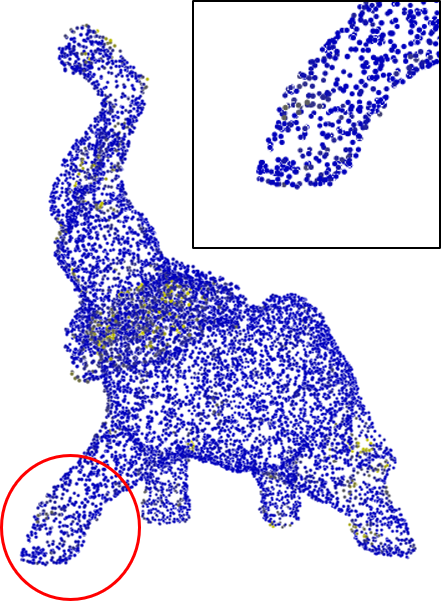}
        \end{minipage}
    }
    \hspace{-0.22cm}
    \subfigure[GPCD++(Ours)]{
        \begin{minipage}[t]{0.19\linewidth}
            \centering
            \includegraphics[width=\linewidth]{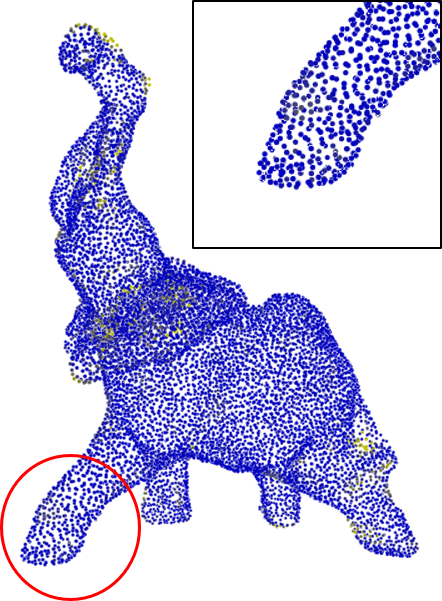}
        \end{minipage}
    }
    \hspace{-0.22cm}
    \subfigure{
        \begin{minipage}[t]{0.03\linewidth}
            \centering
            \includegraphics[width=\linewidth]{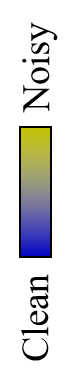}
        \end{minipage}
    }
    \hspace{-0.5cm}
    \caption{\textbf{An overview of our proposed denoising method.} The gradient-based backbone~\cite{luo2021score,chen2021deep} estimates the gradient field and then adopts iterative gradient ascent method for denoising, while our proposed UniNet models local point interaction and achieves distribution uniformity after refinement. }
    \label{fig:teaser}
\end{figure*}

\begin{figure*}[!t]
    \centering
    \subfigure[Uniform]{
            \includegraphics[height=14em]{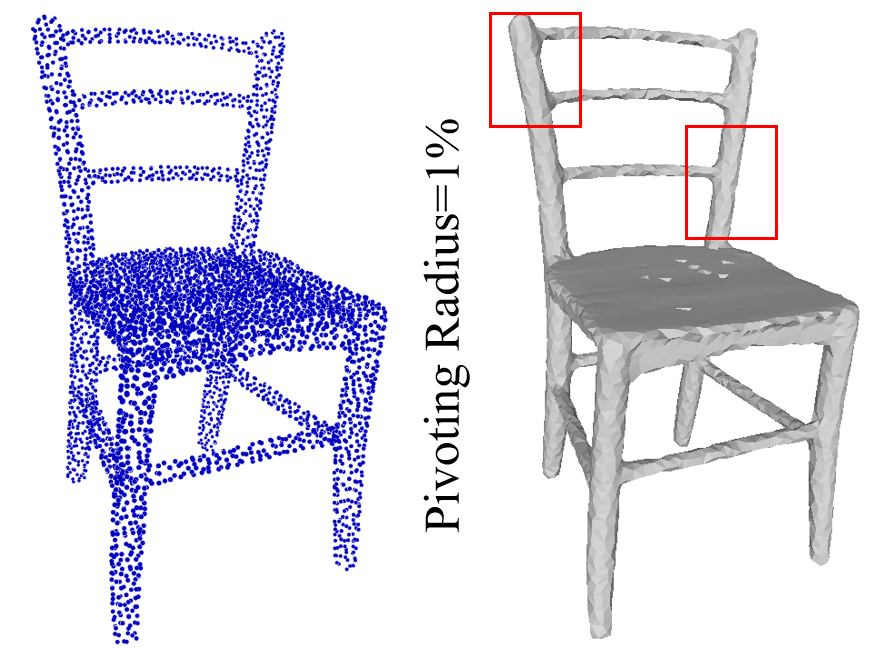}
    }
    \subfigure[Non-Uniform]{
            \includegraphics[height=14em]{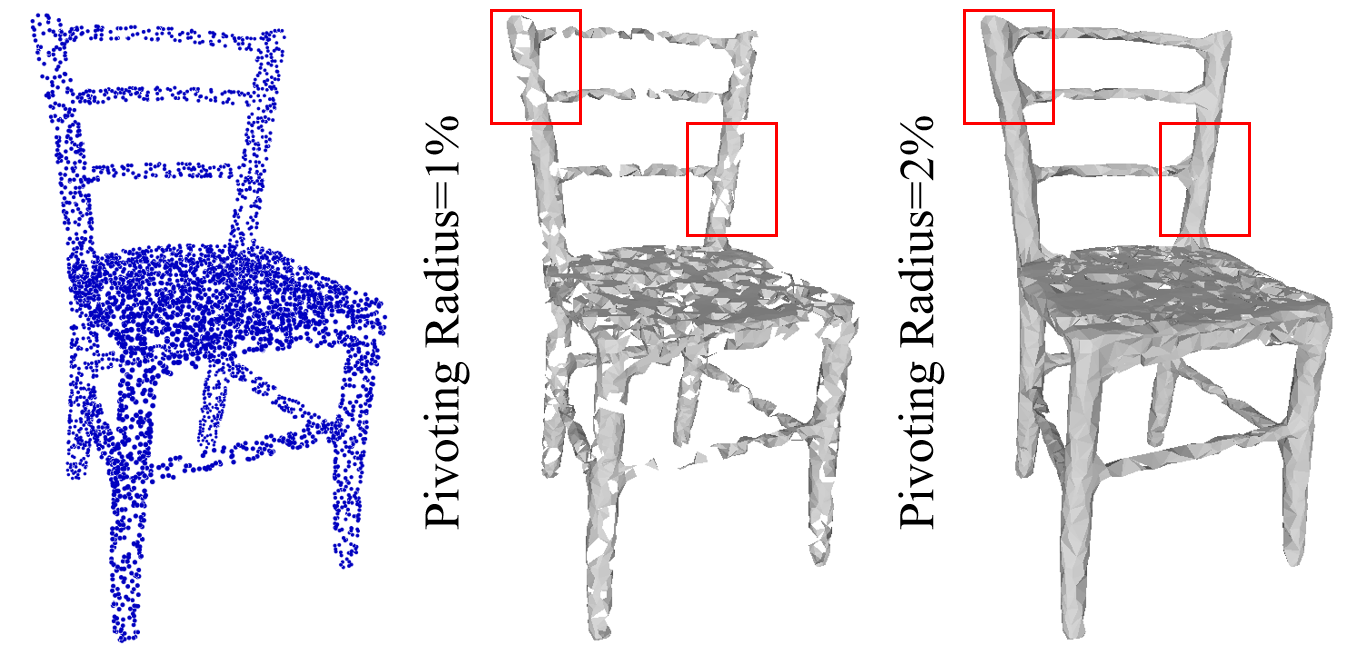}
    }
    \caption{\textbf{Ball pivoting surface reconstruction~\cite{ballpivoting} results of uniform and non-uniform point clouds.} Both point clouds are sampled on the ground truth surface with zero Point-to-Mesh Distance.}
\label{uni}
\end{figure*}

Existing denoising methods mainly focus on aligning noisy points with the underlying surface, and apply Chamfer Distance (CD) and Point-to-Mesh Distance (P2M) to measure the alignment. Despite having achieved promising quantitative results regarding these metrics, previous state-of-the-art methods~\cite{luo2021score,chen2021deep} tend to produce non-uniformly distributed points, as illustrated in Fig.~\ref{fig:teaser}. While being ignored in denoising, the distribution uniformity has been addressed in recent point cloud upsampling methods~\cite{li2021dispu,li2019pugan,yu2018pu}. We argue that uniformity is an essential property for point clouds to better serve downstream tasks like surface reconstruction. As shown in Fig.~\ref{uni}, both point clouds align perfectly with the underlying mesh surface (P2M error is 0), whereas unexpected holes appear in low-density areas of the non-uniform point cloud after surface reconstruction. Moreover, hyper-parameter selection becomes sophisticated for non-uniform point clouds to achieve a balance between filling holes and preserving features.

Inspired by this observation, we conduct an analysis of existing gradient-based methods and discover that the non-uniformity stems from the independence assumption of points.
Points are treated independently without interactions, neglecting contextual information when converging towards the underlying surface, which results in non-uniformly distributed areas like holes and clusters. 
Although regularization is often beneficial to point cloud restoration, previous work~\cite{chen2021deep} demonstrates that introducing conventional regularization into the denoising process has limited improvement on the performance when the noise level is moderate. 
To solve this problem, we approximate the log-likelihood of a noisy point cloud with two terms: a point-wise log-likelihood term, along with a joint log-likelihood term which models interactions among points. We propose GPCD++, a novel denoising framework consisting of a gradient-based denoising backbone and an ultra-lightweight network named UniNet to capture contextual information and ensure uniform distribution. Notably, the backbone network can be adopted from any gradient-based denoising method, such as Score-based Denoising Network~\cite{luo2021score} or Point Set Resampling~\cite{chen2021deep}. 
Specifically, the backbone is employed to model the gradient field of 3D point distribution and predict the direction towards the underlying surface for noisy points.
Following the backbone network, UniNet refines the point cloud for distribution uniformity while keeping points on the underlying surface. It is worth noting that UniNet only consists of two graph convolution layers to model feature interactions among points, which effectively enhances gradient-based methods with minimal computational overhead. 
Extensive results show that GPCD++ achieves competitive or even better denoising performance and significantly improves distribution uniformity on synthetic and real-scanned data. Further experiments demonstrate that higher-quality surfaces can be reconstructed from the point clouds generated by our method.

We summarize our major contributions as follows:
\begin{itemize}
    \item We thoroughly analyze existing gradient-based denoising methods and demonstrate that the distribution non-uniformity stems from the point independence assumption.
    \item We propose GPCD++, a novel denoising framework that leverages a lightweight network named UniNet to effectively achieve distribution uniformity while preserving geometric features and local details.
    \item Extensive experiments show that our network achieves comparable  denoising performance and can significantly improve point distribution uniformity with little computational overhead, which largely benefits downstream tasks like surface reconstruction. 
\end{itemize}

\section{Related Work}

\subsection{Point Cloud Denoising}

Early point cloud denoising methods~\cite{digne2017bilateral,huang2013edge,cazals2005estimating,alexa2001point,avron2010l1,mattei2017point,zaman2017density} are mainly optimization-based, which rely heavily on geometric priors and have difficulty balancing between detail preservation and denoising effectiveness. In recent years, with the advent of point-based neural networks such as PointNet~\cite{qi2017pointnet}, deep-learning-based denoising methods have emerged and achieved promising results. In general, deep-learning-based methods can be roughly divided into three categories: displacement-based~\cite{duan20193d,roveri2018pointpronets,rakotosaona2020pointcleannet,hermosilla2019total,pistilli2020learning,hu2021dynamic}, downsample-upsample-based~\cite{luo2020differentiable}, and gradient-based~\cite{chen2021deep,luo2021score} methods. 
PointCleanNet~\cite{rakotosaona2020pointcleannet} is the pioneer of displacement-based denoising methods, which employs an architecture based on PointNet to estimate the single-step corrective displacement of noisy points. Subsequently, GPDNet~\cite{pistilli2020learning} enhances the denoising network by graph convolutions to exploit the local structure of the neighborhood. However, the estimation of single-step corrective displacement may not be sufficiently accurate thus these methods generally suffer from shrinkage and outliers. 
To address these issues, DMRDenoise~\cite{luo2020differentiable} proposes to explicitly learn the underlying manifold of noisy point clouds from a downsampled subset of points with less noises via a differentiable pooling layer. However, the downsampling stage also discards geometric details and leads to over-smoothing. 
Recently, gradient-based methods~\cite{luo2021score,chen2021deep} formalize point cloud denoising as an iterative process of increasing the log-likehood of each point via estimating the gradient of the underlying distribution. Compared with displacement-based methods which only consider the position of each input point, gradient-based methods model 3D continuous distribution supported by a 2D manifold. Compared with downsample-upsample method, they preserve more informative details at low noise levels. Although they have outperformed previous approaches, gradient-based methods neglect the relationship among points while iterative denoising, which leads to serious distribution non-uniformity.

In this work, we demonstrate that the distribution non-uniformity stems from the point independence assumption, and propose to extend previous gradient-based methods using UniNet. Our method achieves comparable denoising performance and can significantly achieve distribution uniformity, which largely benefits downstream tasks.

\subsection{Distribution Uniformity of Point Clouds}

Although distribution uniformity has not received sufficient attention in point cloud denoising, it has been explored in point cloud upsampling. For upsampling, the generated points should be located on the underlying surface and cover the surface with a uniform distribution. PU-Net~\cite{yu2018pu} proposes to distribute the generated points more uniformly on the object surface via a repulsion loss, which penalizes a point if it is too close to its neighborhoods. Subsequently, PU-GAN~\cite{li2019pugan} adopts a chi-squared model to measure the uniformity and proposes the uniform loss to enhance the output point distribution uniformity. The follow-up Dis-PU~\cite{li2021dispu} disentangles the upsampling task based on its multi-objective nature and adopts a spatial refiner network to generate more uniform dense point sets. All these previous methods demonstrate that high-quality point clouds should not only faithfully locate on the underlying surface, but also cover the surface with a uniform distribution. Regretfully, point clouds produced by existing gradient-based denoisers lack such uniformity, which significantly obstructs downstream tasks such as surface reconstruction as illustrated by our experiments.

\section{Method}

\begin{figure}[!t]
    \centering
    \subfigure[Gradient Field Estimation]{
        \begin{minipage}[t]{0.5\linewidth}
            \centering
            \includegraphics[width=\linewidth]{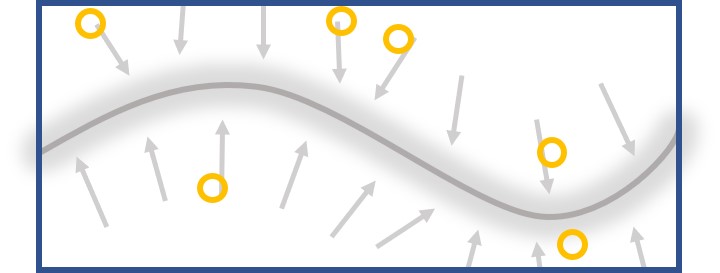}
        \end{minipage}
    }
    \hspace{-0.55cm}
    \subfigure[Gradient Ascent]{
        \begin{minipage}[t]{0.5\linewidth}
            \centering
            \includegraphics[width=\linewidth]{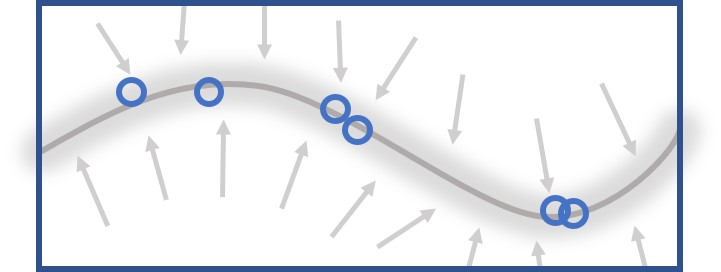}
        \end{minipage}
    }
    \subfigure[Point Interaction]{
        \begin{minipage}[t]{0.5\linewidth}
            \centering
            \includegraphics[width=\linewidth]{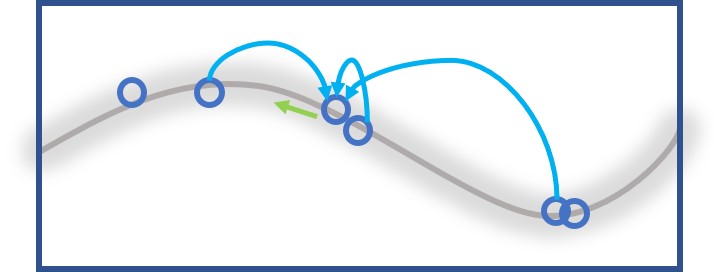}
        \end{minipage}
    }
    \hspace{-0.55cm}
    \subfigure[Refinement]{
        \begin{minipage}[t]{0.5\linewidth}
            \centering
            \includegraphics[width=\linewidth]{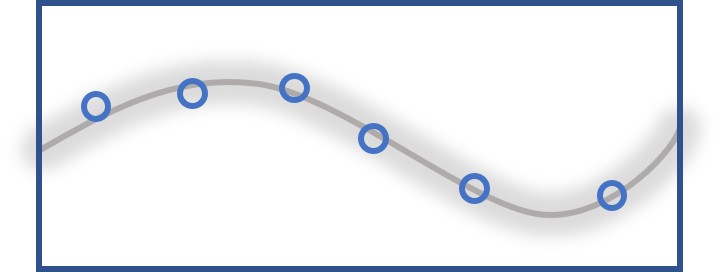}
        \end{minipage}
    }
    \caption{\textbf{An overview of our proposed denoising method.} The gradient-based backbone~\cite{luo2021score,chen2021deep} estimates the gradient field and then adopts iterative gradient ascent method for denoising, while our proposed UniNet models local point interaction and achieves distribution uniformity after refinement.}
    \label{method}
\end{figure}

The most important finding of this paper is that existing gradient-based point cloud denoising methods suffer from the local distribution non-uniformity problem. In this section, we first provide a deep analysis of existing gradient-based methods and give an explanation to this phenomenon (Sec.\ref{method-analysis}). To address this issue, we present GPCD++ (Sec.\ref{method-network}), a novel denoising framework consisting of a gradient-based backbone and an ultra-lightweight network named UniNet. As illustrated in Fig.~\ref{method}, the gradient-based backbone estimates the gradient field for denoising, while UniNet aims to model local point interactions and achieve distribution uniformity using only two graph convolution layers. We briefly introduce our training strategy in Sec.~\ref{method-training}. 


\subsection{Analysis of Gradient-based Denoising Methods}\label{method-analysis}

Given a clean point cloud $\bm{Y}=\{\bm{y}_i\}_{i=1}^n$ containing $n$ points, each point $\bm{y}_i$ can be viewed as a sample from the underlying surface of a 3D object, which can be represented as a 3D distribution $\mathcal P$ supported by a 2D manifold. However, point clouds are often perturbed by noises due to the inherent limitation of depth sensors, which can be formalized as applying a degradation operator $\mathcal{D}$ to the clean point cloud $\bm{Y}$. We represent the noisy point cloud as $\bm{X}=\mathcal{D}(\bm{Y})$ and $\mathcal{D}$ is often modeled as an additive noise term $\bm{N}=\{\bm{n}_i\}_{i=1}^n$ sampled from the noise distribution $\mathcal N$, i.e., $\bm{x}_i = \bm{y}_i + \bm{n}_i$.


Let $q(\cdot)$ denote the probability density function of 3D distribution $\mathcal P$. Gradient-based denoising methods~\cite{luo2021score,chen2021deep} formalize the process of denoising  the input noisy point cloud $\bm{X}$ as finding a transformed point cloud $\widetilde{\bm{X}} =\{\widetilde{\bm{x}}_i\}_{i=1}^n$ that maximize $q(\bm{X})$. To solve this optimal likelihood problem, previous methods view each point as an independent sample from the point distribution. Therefore, the log-likelihood of a point cloud can be written as the summation of point-wise log-likelihoods: 
\begin{align}
    \max_{\bm{X}} \log q(\bm{X}) = \max_{\bm{X}} \sum_{i=1}^n \log q(\bm{x}_i).
\end{align}
Existing gradient-based methods aim at learning the first-order derivative of the log-density function to denoise the input point cloud iteratively. Thus, the denoising process can be written as:
\begin{align}
\label{gpcd_eq}
    {\bm{X}^{(t+1)}} &= \bm{X}^{(t)} + s_t \nabla_{\bm{X}}\sum_{i=1}^n \log q(\bm{x}^{(t)}_i),
\end{align}
where $\bm{X}^{(0)} = \bm{X}$ and $s_t$ is the step size at the $t$-th step.

\begin{figure*}[!t]
\centering
\includegraphics[width=0.9\linewidth]{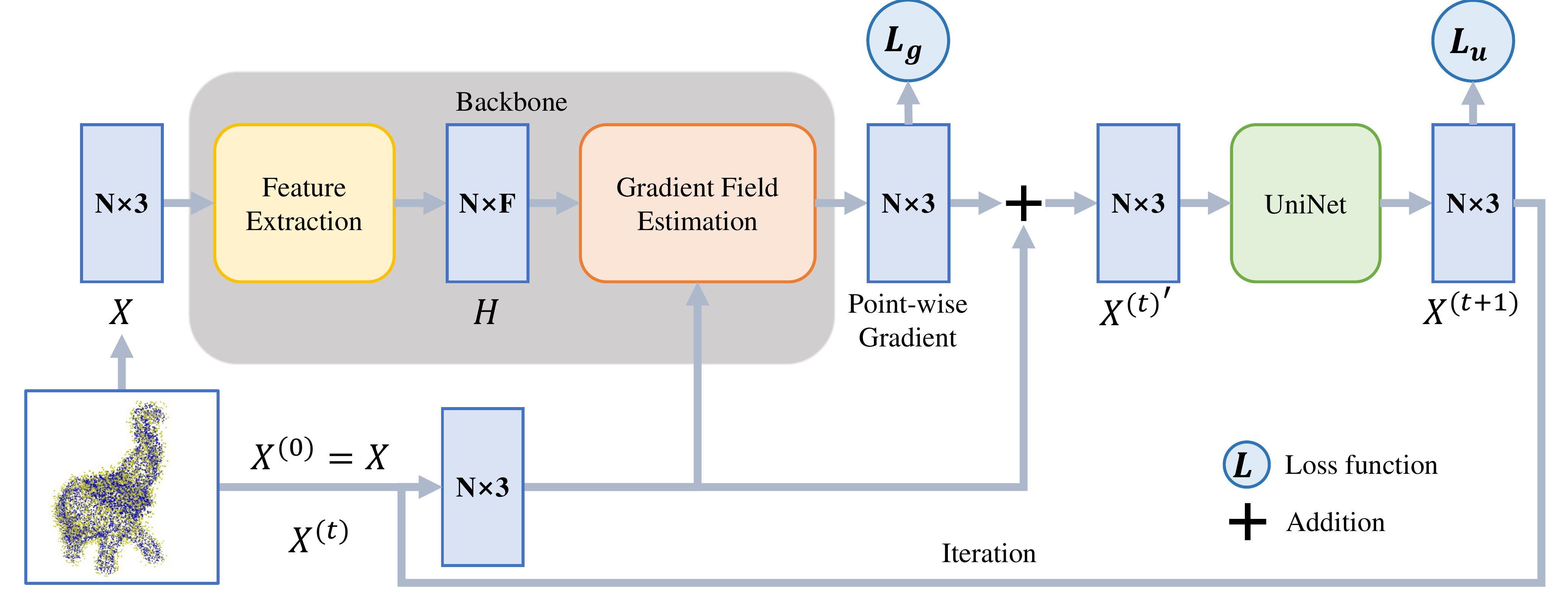}
\caption{\textbf{Illustration of our proposed GPCD++.} GPCD++ consists of a gradient-based denoising backbone and an ultra-lightweight network named UniNet to capture contextual information and ensure uniform distribution. }
\label{fig:gpcd2}
\vspace{-0.5cm}
\end{figure*}

Based on the point independence assumption,the log-likelihood $q(\bm{X})$ is transformed into the summation of $n$ point-wise log-likelihoods $q(\bm{x}_i)$ to reduce the computational complexity. However, high-quality point clouds should not only be faithfully located on the underlying surface of 3D objects, but also cover the surface with a uniform distribution~\cite{li2021dispu,li2019pugan}. As shown in Fig.~\ref{method} (a)(b), neglecting local point interactions leads to severe distribution non-uniformity of the generated point clouds.

On the other hand, due to the lack of ground truth for the gradient field $\nabla_{\bm{X}}\sum_{i=1}^n \log q(\bm{x}_i)$, previous methods use the center $\overline{\bm{y}}$ of $K$ nearest points $\bm{y}_k$ of the noisy point $\bm{x}_i$ to approximate the ground truth gradient for $\bm{x}_i$. However, this approximation results in errors that could not be ignored when the curvature is large or points are sparsely distributed in the local region.
Errors accumulate with iterative gradient ascent, which exacerbates distribution uniformity.

\subsection{Gradient-based Point Cloud Denoising with Uniformity}\label{method-network}

Based on the above analysis, we argue that the core of solving the distribution non-uniformity problem is to model the interactions among points while iteratively denoising the noisy points, as shown in Fig.~\ref{method} (c)(d). However, due to the inherent permutation invariance of point clouds, it is difficult to arrange all points in a specific order and consider previous states when learning the gradient field. To this end, we propose to disentangle the gradient field $\nabla_{\bm{X}} \log q(\bm{X})$ into two parts:
\begin{align}
    \nabla_{\bm{X}} \log q(\bm{X}) = \nabla_{\bm{X}} \sum_{i=1}^n \log q(\bm{x}_i) + \nabla_{\bm{X}}\log g(\bm{x}_1,...,\bm{x}_n),
\end{align}
where $\nabla_{\bm{X}}\log g(\bm{x}_1, ... ,\bm{x}_n)$ is a complement of point-wise gradient that aims at capturing feature interactions among points while denoising.
Compared with Eq.~\ref{gpcd_eq}, the iterative denoising process of GPCD++ can be written as:
\begin{equation}
\begin{split}
    {\bm{X}^{(t+1)}} = \bm{X}^{(t)} + s_t (\nabla_{\bm{X}}\sum_{i=1}^n \log q(\bm{x}^{(t)}_i) \\+ \nabla_{\bm{X}}\log g(\bm{x}^{(t)}_1, ... \bm{x}^{(t)}_n) ),
\end{split}
\end{equation}
where $\bm{X}^{(0)} = \bm{X}$. Previous work~\cite{li2021dispu} shows the challenge for a single network to meet both uniformity and proximity-to-surface at the same time. Therefore, in our implementation, the function $g(\cdot)$ takes one-step denoised points ${\bm{x}_i^{(t)}}' = \bm{x}_i^{(t)}+s_t \nabla_{\bm{x}_i} \log q(\bm{x}^{(t)}_i)$ as input instead of $\bm{x}_i^{(t)}$. The corresponding network is constrained to focus more on distribution uniformity and is easier to train.  

As illustrated in Fig. \ref{fig:gpcd2}, our proposed GPCD++ consists of three main parts, including a feature extraction unit, a gradient field estimation unit, and a lightweight network named UniNet, which is used for modeling feature interactions among points. Given the input noisy point cloud $\bm{X}$, the feature extraction unit produces point-wise features $\bm{H} = \{\bm{h}_i\}_{i=1}^n$ that encode local geometric information. The gradient field estimation unit takes both the output features $\bm{H}$ and the noisy point cloud $\bm{X}^{(t)}$ at the $t$-th step as input and then predicts the point-wise gradient vectors $\nabla_{\bm{x}}\log q(\bm{x}_i)$ pointing to the underlying surface of the 3D objects. Finally, UniNet aims at converging the denoised point cloud ${\bm{X}^{(t)}}' = \bm{X}^{(t)}+s_t \nabla_{\bm{X}}\sum_{i=1}^n \log q(\bm{x}^{(t)}_i)$ towards uniform distribution on the surface. It is worth noting that the denoising backbone can be implemented as any gradient-based denoising network such as Score-based Denosing Network~\cite{luo2021score} or Point Set Resampling~\cite{chen2021deep}. For completeness, we first briefly introduce the gradient-based network architecture, then describe the proposed UniNet which leverages only two graph convolution layers to achieve distribution uniformity while maintaining proximity-to-surface.

\subsubsection{Feature Extraction Unit.} Following previous work~\cite{luo2020differentiable,luo2021score,chen2021deep,yifan2019patch,li2019pugan}, we adopt the feature extractor widely used in denoising and upsampling models as our feature extraction unit, which is a stack of densely connected edge convolution layers~\cite{wang2019dynamic}. More specifically, each point is viewed as a vertex of a graph, which is connected to its $K$ nearest neighbors. The dynamic graph convolution is employed to group the point-wise features and encode both local and nonlocal contextual information without point set subsampling. Then the grouped features are refined by a chain of densely connected MLPs. Finally, the feature extraction unit outputs the point-wise features $\bm{h}_i$ used for gradient field estimation. 

\begin{figure*}[!t]
\centering
\includegraphics[width=0.90\linewidth]{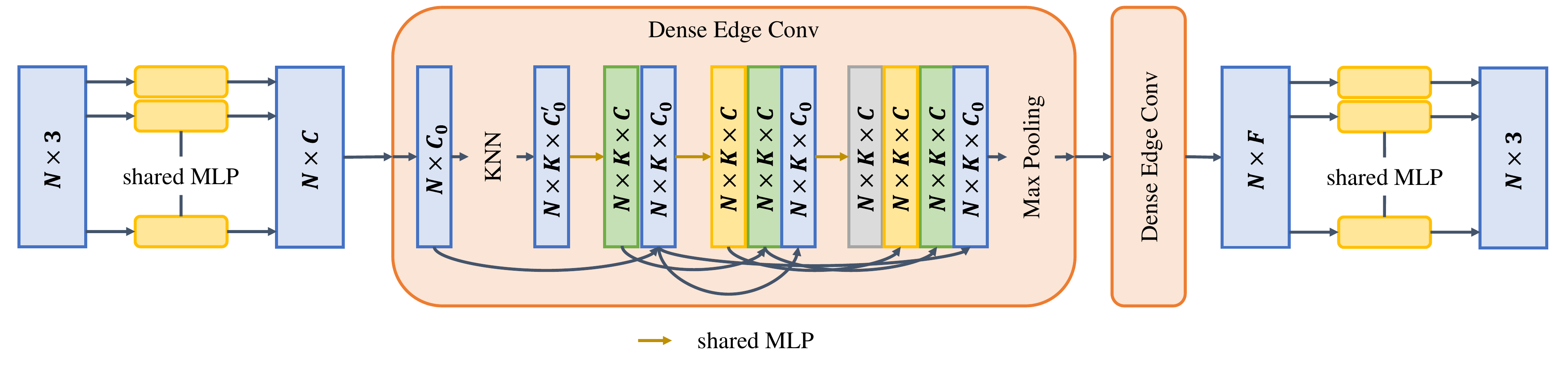}
\caption{\textbf{An overview of our proposed UniNet.} "mlp" stands for multi-layer perceptons. UniNet is an ultra-lightweight network consisting of two graph convolution layers.}
\label{UniNet}
\end{figure*}

\subsubsection{Gradient Field Estimation Unit.} The gradient field estimation unit aims at estimating the gradient ascent direction for each point $\bm{x}_i^{(t)}$ at the $t$-th step. Gradient-based methods sample $K$ nearest neighbors $\bm{x}_j$ from the input noisy point cloud as well as their geometric features $\bm{h}_j$, then aggregate information to predict the gradient ascent direction. Score-based Denoising Network~\cite{luo2021score} proposes to use late fusion in the Euclidean space, while Point Set Resampling~\cite{chen2021deep} adopts a cosine annealing module to fuse information in the feature space.

\subsubsection{UniNet.} Our proposed GPCD++ adopts an ultra-lightweight network named UniNet to model feature interactions among points. Given the denoised point cloud ${\bm{X}^{(t)}}'$ from the gradient field estimation unit, UniNet aims at learning the gradient of the joint log-likelihood term $\nabla_{\bm{X}}\log g({\bm{x}^{(t)}_1}', ... ,{\bm{x}^{(t)}_n}')$ to improve both proximity-to-surface and distribution uniformity. Notably, compared with point-based neural networks on classification or segmentation, UniNet only needs to focus on local geometry and point distribution. Therefore, we propose to adopt only two graph convolutions to capture local contextual information for both computational efficiency and refinement effectiveness. In contrast, most previous operators~\cite{pointnet2,wang2019dynamic} on point clouds are designed for semantic tasks such as classification and segmentation, which have difficulty in capturing local point distribution when the depth of neural networks is low. To enhance the representational power of graph convolution, we leverage dense connections following previous denoising and upsampling models~\cite{li2019pugan,luo2020differentiable,yifan2019patch,chen2021deep}.

As illustrated in Fig.~\ref{UniNet}, the proposed UniNet consists of three parts, including two shared multi-layer perceptrons (MLP) at each point and a graph neural network with two stacked graph convolution layers. Specifically, given the input point-wise features $\bm{F}=\{\bm{f}_i\}_{i=1}^N$ output by shared MLP, we first group the features using kNN search based on Euclidean distance instead of feature similarity, considering that long-range and non-local information is not essential to our task.
After that, we refine each grouped feature using densely connected MLPs introduced in ~\cite{yifan2019patch}, and finally adopt a max-pooling layer as a symmetric function to aggregate information from the local region.

\subsection{Network Training}\label{method-training}

If we train GPCD++ from scratch, the low-quality output point clouds of the denoising backbone lead to an unstable training process of UniNet at the early training stage.
Thus, we pretrain the feature extraction unit and the gradient field estimation unit on the same dataset using the loss function $\mathcal{L}_g$, which follows previously proposed training objective~\cite{luo2021score,chen2021deep}, and then freeze them while training UniNet with the loss function $\mathcal{L}_u$.

$\mathcal{L}_u$ measures the difference between the output point cloud and the ground truth point cloud. Chamfer Distance (CD) and Earth Mover's Distance (EMD) are two permutation-invariant loss functions widely used in various tasks on point clouds (e.g., completion, upsampling, denoising). Supposing $\bm{X}''=\{\bm{x}''_i\}_{i=1}^N$ and $\bm{Y}=\{\bm{y}_i\}_{i=1}^N$ denotes the output and ground truth point cloud, the loss functions are defined as:
\begin{gather}
    \begin{split}
        \text{CD}(\bm{X}'', \bm{Y}) = \frac{1}{N}\sum_{\bm{x}\in \bm{X}''}\min_{\bm{y}\in \bm{Y}}||\bm{x}-\bm{y}||^2_2 \\ + \frac{1}{N}\sum_{\bm{y}\in \bm{Y}}\min_{\bm{x}\in \bm{X}''}||\bm{x}-\bm{y}||^2_2
    \end{split} \\
    \text{EMD}(\bm{X}'', \bm{Y}) = \min_{\phi:X''\rightarrow Y} \frac{1}{N}\sum_{\bm{x}\in X''}||\bm{x}-\phi(\bm{x})||^2_2
\end{gather}
Here CD directly calculates the average closest point distance between two point clouds $\bm{X}''$ and $\bm{Y}$, while EMD finds a bijection $\phi :\bm{X}''\rightarrow \bm{Y}$ which minimizes the average distance between corresponding points.
Compared with CD, EMD can provide stronger supervision to train UniNet, so we use EMD to regularize the output point clouds for point cloud denoising. 
Notably, our method does not require other loss terms, since the network trained with EMD is already able to refine the point clouds for proximity-to-surface and distribution uniformity at the same time.

\section{Experiments}


\subsection{Dataset}

Following previous works~\cite{luo2021score,chen2021deep}, we collect 20 meshes from the training subset of PU-Net and use Poisson disk sampling to generate ground truth point clouds ranging from 10k points to 50k points. After normalization into the unit sphere, point clouds are perturbed by Gaussian noise with zero mean and a standard deviation from 0.5\% to 2\% of the bounding ball radius. Each training point cloud is split into patches before fed into the network to reduce memory use. In training, we use random sampling to sample patches(patch size is 1K).

For testing, we compare our model with state-of-the-art methods on the testing subset of PU-Net~\cite{yu2018pu} and PointCleanNet~\cite{rakotosaona2020pointcleannet} datasets, which contain 20 meshes and 10 meshes, respectively. Three metrics are employed to evaluate point cloud denoising methods, including Chamfer distance, Point-to-Mesh Distance for denoising effectiveness, and modified uniformity metric~\cite{li2019pugan} for point uniformity.
We normalize the denoised point clouds into a unit sphere before computing the metrics. For uniformity metric, we adopt the farthest sampling to pick $M$ seed points, and then crop a point set $S_j$ at each seed using ball query. The uniformity metric can be written as: 
\begin{align}
    \label{uniformity_metric}
    \text{Uni} =\frac{1}{M} \sum_{j=1}^M U_\text{imbalance} (S_j) \times U_\text{clutter}(S_j),
\end{align}
where $U_\text{imbalance} (S_j) = \frac{(|S_j|-\widehat{n})^2}{\widehat{n}}$ denotes the deviation of $|S_j|$ from $\widehat{n}$. Here $\widehat{n}$ is the expected number of points in $S_j$. $U_\text{clutter}(S_j)$ measures the deviation of each point's distance to its nearest neighbor, denoted by $d_{j,k}$, and is defined as $U_\text{clutter}(S_j) = \frac{1}{|S_j|} \sum_{k=1}^{|S_j|} \frac{(d_{j,k}-\widehat{d})^2}{\widehat{d}}$. $\widehat{d}$ is the expected distance between a point and its nearest neighbor. Compared with original uniform loss, our modified one leverages the ground truth point cloud to compute the expected distance $\widehat{d}$ and expected number $\widehat{n}$, which provides a more accurate estimation, especially for objects with complex geometric shape.

\subsection{Noise Models}

To demonstrate the generalization, we test our proposed model with point clouds perturbed by several types of noise, including isotropic Gaussian noise, Laplace noise, discrete noise, anisotropic Gaussian noise, uni-directional Gaussian noise, uniform noise, simulated sensor noise and real noise. The following are the setting of corresponding hyperparameters, where the scale parameter $s$ is set to 1\%, 2\% and 3\% of the bounding sphere radius to simulate different levels of noise.

\begin{itemize}
\item Isotropic Gaussian noise.
    \begin{align}
        p(x;s) = \frac{1}{\sqrt{2\pi}s}e^{-\frac{x^2}{2s^2}} \nonumber
    \end{align}
\item Laplace noise.
    \begin{align}
        p(x;s) = \frac{1}{2s}e^{-\frac{|x|}{s}} \nonumber
    \end{align}
\item Discrete noise.
    \begin{align}
        p(x;s) = \left\{\begin{matrix}
        0.1 & (\pm s,0,0),(0,\pm s,0),(\pm 0,0,\pm s)\\ 
        0.4 & (0,0,0)\\ 
        0 & \text{otherwise}
        \end{matrix}\right. \nonumber
    \end{align}
\item Anisotropic Gaussian noise.
    \begin{align}
        x &\sim \mathcal{N}(0,\Sigma) \nonumber\\
        \text{where} \;\; \Sigma &= s^2 \times \begin{bmatrix}
        1 & -\frac{1}{2} & -\frac{1}{4}\\ 
        -\frac{1}{2} & 1 & -\frac{1}{4} \\ 
        -\frac{1}{4} & -\frac{1}{4} & 1
        \end{bmatrix} \nonumber
    \end{align}
\item Uni-directional Gaussian noise. We only perturb the point clouds along x-axis using Gaussian noise. 
\item Uniform noise.
    \begin{align}
        p(x;s) = \left\{\begin{matrix}
        \frac{3}{4\pi s^3} & ||x||_2\leq s\\ 
        0 & \text{otherwise}
        \end{matrix}\right. \nonumber
    \end{align}
\item Simulated sensor noise. We adopt the Blensor~\cite{gschwandtner2011blensor} simulation package to simulate realistic noise.
\item Real noise. We test our method on the real-world dataset Paris-rue-Madame~\cite{paris-rue-madame}, which is obtained from real laser scanner.
    
\end{itemize}

\subsection{Implementation Details}

In the training phase, we perform data augmentation on the fly by randomly scaling and rotating the input point clouds.
We adopt Adam optimizer with an initial learning rate $2e-4$ to train UniNet. 
The learning rate is multiplied by $0.8$ on epoch 30, 60 and 90. 

In the testing phase, for uniformity metric, the number of seed points $M$ is set as $rN$, where $N$ is the number of points and $r$ is $0.05$. We crop each set $S_j$ with radius $r_d=\sqrt{p}$ for each $p\in\{$ 0.4\%, 0.6\%, 0.8\%, 1.0\% $\}$, and compute Eq.~\ref{uniformity_metric} four times, then sum up the results. 

\setlength{\tabcolsep}{3pt}
\begin{table*}
\begin{center}
\caption{\textbf{Comparison among competitive denoising algorithms under isotropic Gaussian noise.} CD and P2M are multiplied by $10^4$. Our method surpasses previous state-of-the-art methods on both benchmark datasets.}
\label{benchmark}
\begin{tabular}{c|c|cccccc|cccccc}
\hline
\multicolumn{2}{c|}{\# Points} & \multicolumn{6}{c|}{10K(Sparse)} & \multicolumn{6}{c}{50K(Dense)} \\
\multicolumn{2}{c|}{Noise} & \multicolumn{2}{c}{1\%} & \multicolumn{2}{c}{2\%} & \multicolumn{2}{c|}{3\%} & \multicolumn{2}{c}{1\%} & \multicolumn{2}{c}{2\%} & \multicolumn{2}{c}{3\%}\\
Dataset & Model &  CD & P2M&  CD & P2M&  CD & P2M&  CD & P2M&  CD & P2M&  CD & P2M\\
\hline
\multirow{10}{*}{PU} & Bilateral & 3.646 & 1.342 & 5.007 & 2.018 & 6.998 & 3.557 & 0.877 & 0.234 & 2.376 & 1.389 & 6.304 & 4.730 \\
 & Jet & 2.712 & 0.613 & 4.155 & 1.347 & 6.262 & 2.921 & 0.851 & 0.207 & 2.432 & 1.403 & 5.788 & 4.267 \\
 & MRPCA & 2.972 & 0.922 & 3.728 & 1.117 & 5.009 & 1.963 & 0.669 & 0.099 & 2.008 & 1.033 & 5.775 & 4.081 \\
 & GLR & 2.959 & 1.052 & 3.773 & 1.306 & 4.909 & 2.114 & 0.696 & 0.161 & 1.587 & 0.830 & 3.839 & 2.707 \\
 \cline{2-14}   
 & PCN & 3.515 & 1.148 & 7.467 & 3.965 & 13.067 & 8.737 & 1.049 & 0.346 & 1.447 & 0.608 & 2.289 & 1.285 \\
 & GPDNet & 3.780 & 1.337 & 8.007 & 4.426 & 13.482 & 9.114 & 1.913 & 1.037 & 5.021 & 3.736 & 9.705 & 7.998 \\
 & DMR & 4.482 & 1.722 & 4.982 & 2.115 & 5.892 & 2.846 & 1.162 & 0.469 & 1.566 & 0.800 & 2.432 & 1.528 \\
 & Score-based & 2.521 & 0.463 & 3.686 & 1.074 & 4.708 & 1.942 & 0.716 & 0.150 & 1.288 & 0.566 & 1.928 & 1.041 \\
 & PSR & 2.353 & 0.306 & 3.350 & 0.734 & 4.075 & 1.242 & 0.649 & 0.076 & 0.997 & \textbf{0.296} & 1.344 & \textbf{0.531} \\
 \cline{2-14}
 & GPCD++(Score) & 2.312 & 0.506 & 3.215 & 0.999 & 4.119 & 1.829 & 0.639 & 0.159 & 1.124 & 0.503 & 1.910 & 1.086\\
 & GPCD++(PSR) & \textbf{1.881} & \textbf{0.251} & \textbf{2.728} & \textbf{0.654} & \textbf{3.433} & \textbf{1.161} & \textbf{0.505} & \textbf{0.073} & \textbf{0.852} & 0.303 & \textbf{1.198} & 0.534\\
 \hline
 \multirow{10}{*}{PC} & Bilateral & 4.320 & 1.351 & 6.171 & 1.646 & 8.295 & 2.392 & 1.172 & 0.198 & 2.478 & 0.634 & 6.077 & 2.189 \\
 & Jet & 3.032 & 0.830 & 5.298 & 1.372 & 7.650 & 2.227 & 1.091 & 0.180 & 2.582 & 0.700 & 5.787 & 2.144 \\
 & MRPCA & 3.323 & 0.931 & 4.874 & 1.178 & 6.502 & 1.676 & 0.966 & 0.140 & 2.153 & 0.478 & 5.570 & 1.976 \\
 & GLR & 3.399 & 0.956 & 5.274 & 1.146 & 7.249 & 1.674 & 0.964 & 0.134 & 2.015 & 0.417 & 4.488 & 1.306\\
 \cline{2-14}
 & PCN & 3.847 & 1.221 & 8.752 & 3.043 & 14.525 & 5.873 & 1.293 & 0.289 & 1.913 & 0.505 & 3.249 & 1.076 \\
 & GPDNet & 5.470 & 1.973 & 10.006 & 3.650 & 15.521 & 6.353 & 5.310 & 1.716 & 7.709 & 2.859 & 11.941 & 5.130 \\
 & DMR & 6.602 & 2.152 & 7.145 & 2.237 & 8.087 & 2.487 & 1.566 & 0.350 & 2.009 & 0.485 & 2.993 & 0.859 \\
 & Score-based & 3.369 & 0.830 & 5.132 & 1.195 & 6.776 & 1.941 & 1.066 & 0.177 & 1.659 & 0.354 & 2.494 & 0.657\\
 & PSR & 2.873 & 0.783 & 4.757 & 1.118 & 6.031 & 1.619 & 1.010 & 0.146 & 1.515 & 0.340 & 2.093 & 0.573\\
 \cline{2-14}
 & GPCD++(Score) & 3.379 & 0.787 & 4.707 & 1.084 & 5.972 & 1.621 & 1.026 & 0.191 & 1.526 & 0.337 & 2.497 & 0.693\\
 & GPCD++(PSR) & \textbf{2.813} & \textbf{0.759} & \textbf{4.195} & \textbf{0.893} & \textbf{5.385} & \textbf{1.333} & \textbf{0.857} & \textbf{0.132} & \textbf{1.344} & \textbf{0.331} & \textbf{1.920} & \textbf{0.530}\\
 \hline
\end{tabular}
\end{center}
\end{table*}

\subsection{Quantitative Results} Following previous works~\cite{luo2021score,chen2021deep}, we first perturb the point clouds with isotropic Gaussian noise to compare our models with state-of-the-art point cloud denoising methods. They include both optimization-based methods such as bilateral filtering (Bilateral) ~\cite{digne2017bilateral}, jet fitting (Jet) ~\cite{cazals2005estimating}, MRPCA~\cite{mattei2017point}, GLR ~\cite{zeng20193d}, and learning-based methods including Point Clean Net (PCN) ~\cite{rakotosaona2020pointcleannet}, GPDNet~\cite{pistilli2020learning}, DMRDenoise (DMR) ~\cite{luo2020differentiable}, Score-based Denoising Network (Score-based) ~\cite{luo2021score}, Point Set Resampling via Gradient Field (PSR) ~\cite{chen2021deep}. Notably, the last two methods are gradient-based methods, both of which can be employed as the backbone in our proposed framework GPCD++, thus we implement two versions of GPCD++, named GPCD++(Score) and GPCD++(PSR), to make a comparison.

The experimental results are summarized in Tab.~\ref{benchmark}, where we adopt the results of previous works reported in Point Set Resampling~\cite{chen2021deep} following the same evaluation protocol. We observe noticeable performance gains from not using UniNet in GPCD++(Score) in the first few steps of gradient descent iteration, thus we activate UniNet after 20 steps in GPCD++(Score). It is worth noting that our proposed GPCD++(PSR) surpasses previous state-of-the-art methods on both benchmark datasets, with an obvious improvement on CD and P2M metrics. Compared with Score-based Denoising Network and Point Set Resampling, the corresponding GPCD++ versions almost achieve a consistent improvement on Chamfer Distance while taking both distribution uniformity and proximity-to-surface into consideration, which demonstrates the effectiveness of UniNet. We also note that the performance of GPCD++ on P2M metrics is slightly worse than that of its backbone network in a few cases. The reason is that EMD adopts a point-to-point bijection function $\phi$ to regularize each point and sometimes the correspondence is incorrect. UniNet moves the point $x$ away from the underlying surface and slightly degrades the performance on P2M metrics. 

To explore whether our proposed UniNet improves the distribution uniformity of the output point clouds, we compare GPCD++ with its corresponding backbone on PU dataset using uniformity metric. As shown in Tab.~\ref{uniformity}, GPCD++ significantly outperforms its backbone network under different levels of noise on uniformity metric, while slightly affecting denoising in a few cases. 
With regard to Chamfer Distance, GPCD++ can generate point clouds of higher quality for downstream tasks. 
\setlength{\tabcolsep}{1.2pt}
\begin{table}
\begin{center}
\caption{\textbf{Comparison with gradient-based methods under isotropic Gaussian noise on PU dataset.} CD is multiplied by $10^4$, P2M is multiplied by $10^4$ and Uni is multiplied by $10^3$. Our proposed GPCD++ substantially improves distribution uniformity. }
\label{uniformity}
\begin{tabular}{c|ccc|ccc|ccc}
\hline
\multicolumn{1}{c}{\# Points} & \multicolumn{9}{c}{10K(Sparse)}\\
\multicolumn{1}{c|}{Noise} & \multicolumn{3}{c}{1\%} & \multicolumn{3}{c}{2\%} & \multicolumn{3}{c}{3\%}\\
Model & CD &P2M& Uni & CD & P2M& Uni& CD & P2M& Uni \\
\hline
 Score-based & 2.521 & \textbf{0.463} & 6.493 & 3.686 & 1.074 & 9.666 & 4.708 & 1.942 & 24.040 \\
 GPCD++(Score) & \textbf{2.312} & 0.506 & \textbf{1.445} & \textbf{3.215} & \textbf{0.999} & \textbf{3.287} & \textbf{4.119} & \textbf{1.829} & \textbf{8.414}\\
\hline
 PSR & 2.353 & 0.306 & 3.458 & 3.350 & 1.453 & 7.559 & 4.075 & 1.481 & 10.795 \\
 GPCD++(PSR) & \textbf{1.881} & \textbf{0.251} & \textbf{0.239} &\textbf{2.728} & \textbf{0.654} & \textbf{0.770} & \textbf{3.433} & \textbf{1.161} & \textbf{1.671} \\
\hline
\multicolumn{1}{c|}{\# Points} & \multicolumn{9}{c}{50K(Dense)}\\
\multicolumn{1}{c|}{Noise} & \multicolumn{3}{c}{1\%} & \multicolumn{3}{c}{2\%} & \multicolumn{3}{c}{3\%}\\
Model & CD &P2M& Uni & CD & P2M& Uni& CD & P2M& Uni \\
\hline
 Score-based & 0.716 & \textbf{0.150} & 6.591 & 1.288 & 0.566 & 26.075 & \textbf{1.928} & \textbf{1.041} & 47.393 \\
 GPCD++(Score) & \textbf{0.639} & 0.159 & \textbf{1.767} & \textbf{1.124} & \textbf{0.503} & \textbf{17.013} & 1.910 & 1.086 & \textbf{36.496} \\
\hline
 PSR & 0.649 &  0.076 & 1.602 & 0.997 & \textbf{0.296} & 3.246 & 1.344 & \textbf{0.531} & 5.960 \\
 GPCD++(PSR) & \textbf{0.505} & \textbf{0.073} & \textbf{0.139} & \textbf{0.852} & {0.303} & \textbf{0.750} & \textbf{1.198} & {0.534} & \textbf{2.082}\\
\hline
\end{tabular} 
\end{center}
\vspace{-0.6cm}
\end{table}

Tab.~\ref{noises} illustrates the quantitative results of our proposed GPCD++ under various types of noise, including Laplace noise, discrete noise, anisotropic Gaussian noise, uni-directional Gaussian noise and uniform noise. We take the results reported by Point Set Resampling~\cite{chen2021deep} with the identical evaluation protocol. It is worth noting that we only train our model using Gaussian noise without further fine-tuning. The experimental results demonstrate our model has better generalization capability than previous methods under unseen noise models.

\setlength{\tabcolsep}{1.5pt}
\begin{table}
\begin{center}
\caption{\textbf{Comparison among competitive denoising methods under various types of noise on PU dataset.} CD and P2M are multiplied by $10^4$.}
\label{noises}
\begin{tabular}{c|c|cccccc}
\hline
\multicolumn{2}{c|}{\# Points} & \multicolumn{6}{c}{10K(Sparse)} \\
\multicolumn{2}{c|}{Noise} & \multicolumn{2}{c}{1\%} & \multicolumn{2}{c}{2\%} & \multicolumn{2}{c}{3\%}\\
Type & Model &  CD & P2M&  CD & P2M&  CD & P2M\\
\hline
\multirow{7}{*}{Laplace} & MRPCA & 2.950 & 0.724 & 4.216 & 1.428 & 7.951 & 4.441 \\
 & GLR & 3.223 & 1.121 & 4.751 & 2.090 & 7.977 & 4.773 \\
 & PCN & 4.616 & 1.940 & 11.082 & 7.218 & 20.981 & 15.922\\
 & Score-based & 2.915 & 0.674 & 4.601 & 1.799 & 6.332 & 3.271 \\
 & PSR & 2.663 & 0.450 & 3.790 & 1.067 & 5.110 & 2.017 \\
 \cline{2-8}
 & GPCD++(Score) & 2.657 & 0.661 & 3.903 & 1.433 & 5.215 & 2.615 \\
 & GPCD++(PSR) & \textbf{2.218} & \textbf{0.403} & \textbf{3.152} & \textbf{0.976} & \textbf{4.514} & \textbf{1.970}\\
 \hline
 
 \multirow{7}{*}{Discrete} & MRPCA & 1.522 & 0.629 & 2.353 & 0.674 & 2.607 & 0.743 \\
 & GLR & 1.838 & 1.014 & 2.665 & 1.047 & 2.952 & 1.116 \\
 & PCN & 1.177 & 0.307 & 2.870 & 0.871 & 4.028 & 1.674\\
 & Score-based & 1.249 & 0.251 & 2.177 & 0.416 & 2.653 & 0.653 \\
 & PSR & 1.021 & 0.163 & 1.921 & 0.268 & 2.274 & 0.431 \\
 \cline{2-8}
 & GPCD++(Score) & 1.179 & 0.281 & 2.050 & 0.437 & 2.888 & 0.986\\
 & GPCD++(PSR) & \textbf{0.765} & \textbf{0.138} & \textbf{1.645} & \textbf{0.225} & \textbf{2.081} & \textbf{0.379}\\
 \hline
 
 \multirow{7}{*}{Anisotropic} & MRPCA & 2.676 & 0.689 & 3.605 & 1.007 & 5.108 & 2.081 \\
 & GLR & 2.910 & 1.048 & 3.779 & 1.332 & 4.975 & 2.195 \\
 & PCN & 3.432 & 1.129 & 7.393 & 3.940 & 12.952 & 8.654\\
 & Score-based & 2.470 & 0.456 & 3.682 & 1.084 & 4.776 & 2.000 \\
 & PSR & 2.305 & 0.308 & 3.345 & 0.758 & 4.152 & 1.350 \\
 \cline{2-8}
 & GPCD++(Score) & 2.284 & 0.503 & 3.315 & 1.017 & 4.323 & 1.915\\
 & GPCD++(PSR) & \textbf{1.846} & \textbf{0.254} & \textbf{2.711} & \textbf{0.652} & \textbf{3.524} & \textbf{1.253}\\
 \hline

 \multirow{7}{*}{Uni-dir} & MRPCA & 1.712 & 0.646 & 2.564 & 0.767 & 3.237 & 1.063 \\
 & GLR & 2.033 & 1.026 & 2.837 & 1.139 & 3.472 & 1.434 \\
 & PCN& 1.530 & 0.432 & 3.466 & 1.360 & 5.638 & 2.914\\
 & Score-based & 1.442 & 0.279 & 2.412 & 0.543 & 3.391 & 1.108 \\
 & PSR & 1.256 & 0.196 & 2.196 & 0.386 & 2.862 & 0.686 \\
 \cline{2-8}
 & GPCD++(Score) & 1.402 & 0.313 & 2.235 & 0.529 & 3.112 & 1.111\\
 & GPCD++(PSR) & \textbf{1.019} & \textbf{0.171} & \textbf{1.855} & \textbf{0.332} & \textbf{2.451} & \textbf{0.618}\\
 \hline

 \multirow{7}{*}{Uniform} & MRPCA & 1.555 & 0.633 & 2.754 & 0.684 & 3.229 & 0.765 \\
 & GLR & 1.850 & 1.015 & 2.948 & 1.052 & 3.400 & 1.109 \\
 & PCN& 1.205 & 0.337 & 3.378 & 1.018 & 5.044 & 1.995\\
 & Score-based & 1.277 & 0.248 & 2.467 & 0.418 & 3.079 & 0.654 \\
 & PSR & 1.056 & 0.164 & 2.348 & 0.275 & 2.916 & 0.443 \\
 \cline{2-8}
 & GPCD++(Score) & 1.192 & 0.286 & 2.223 & 0.435 & 3.028 & 0.981\\
 & GPCD++(PSR) & \textbf{0.762} & \textbf{0.136} & \textbf{1.800} & \textbf{0.221} & \textbf{2.302} & \textbf{0.353}\\
 \hline
\end{tabular}
\end{center}
\end{table}

\begin{table}[!t]
\begin{center}
\caption{\textbf{Comparison of denoising simulated lidar data on PU dataset.} CD and P2M are multiplied by $10^4$. Noise level is 1\%.}
\label{comp_real}
\setlength{\tabcolsep}{3pt}
\begin{tabular}{c|cc|c|c}
\hline
\multicolumn{1}{c|}{\# Points} & \multicolumn{4}{c}{Variable(12K$\sim$72K)} \\
Model & CD &P2M & Params(M) & Speed(s/pcd) \\
\hline
 Score-based & 3.942 & 1.823 & 0.187 & 9.05\\
 GPCD++(Score) & \textbf{3.772} & \textbf{1.743} & 0.207(+10.7\%) & 9.67(+6.9\%)\\
\hline
 PSR & 3.864 & 1.644 & 0.267 & 5.49\\
 GPCD++(PSR) & \textbf{3.604} & \textbf{1.599} & 0.286(+7.1\%) & 6.40(+16.6\%)\\
\hline
\end{tabular} 
\vspace{-0.5cm}

\label{realsensor}
\end{center}
\end{table}

We also test on simulated noise (with ground truth mesh) of a real sensor provided by Blensor~\cite{gschwandtner2011blensor} package and the results are shown in Tab.~\ref{realsensor}. 
GPCD++(Score) and GPCD++(PSR) have only 10.7\% and 7.1\% more parameters, respectively, than the corresponding backbones. For inference time, GPCD++(PSR) has a consistent and significant improvement over Point Set Resampling with only 16.6\% more inference time on average. Notably, GPCD++(Score) only adds neglectable inference time compared with Score-based Denoising Network, a patch-based denoising method that also requires time to split the noisy input into patches. In summary, our method outperforms previous methods with minimal cost regarding inference time and model size. 

In particular, another alternative solution is to run sequentially the gradient-based denoiser and then a method to achieve distribution uniformity. To evaluate in this setting, we choose PSR as the denoiser and adopt Graph Laplacian (GL)~\cite{luo2018uniformization} to resample the denoised point cloud. As shown in Tab. \ref{seq}, this ``denoise then resample" baseline slightly improves Chamfer Distance, but brings additional noise (see P2M) and significant computational overhead.

\begin{table}[!t]
\begin{center}
\caption{\textbf{Comparison with ``denoise then resample" under isotropic Gaussian noise on PU dataset.}
 CD and P2M are multiplied by $10^4$.
}

\setlength{\tabcolsep}{2pt}
\begin{tabular}{c|cc|cc|cc|c}
\hline

\multicolumn{1}{c|}{\# Points} & \multicolumn{7}{c}{10K} \\
\multicolumn{1}{c|}{Noise} & \multicolumn{2}{c}{1\%} & \multicolumn{2}{c}{2\%} & \multicolumn{2}{c|}{3\%} & \\
Model & CD &P2M & CD & P2M& CD & P2M & Speed(s/pcd)\\
\hline
 PSR & 2.353 & 0.306  & 3.350 & 0.734 & 4.075 & 1.242 & 1.61 \\
 PSR+GL & 2.562 & 0.331 & 3.122 & 0.748 & 3.907 & 1.286 & 7.67(+6.06)\\
 GPCD++(PSR) & \textbf{1.881} & \textbf{0.251} &\textbf{2.728} & \textbf{0.654} & \textbf{3.433} & \textbf{1.161} & 1.85(+0.24) \\
\hline
\end{tabular} 
\vspace{-0.5cm}

\label{seq}
\end{center}
\end{table}



\subsection{Qualitative Results}

\begin{figure*}[!t]
\centering
\includegraphics[width=0.9\linewidth]{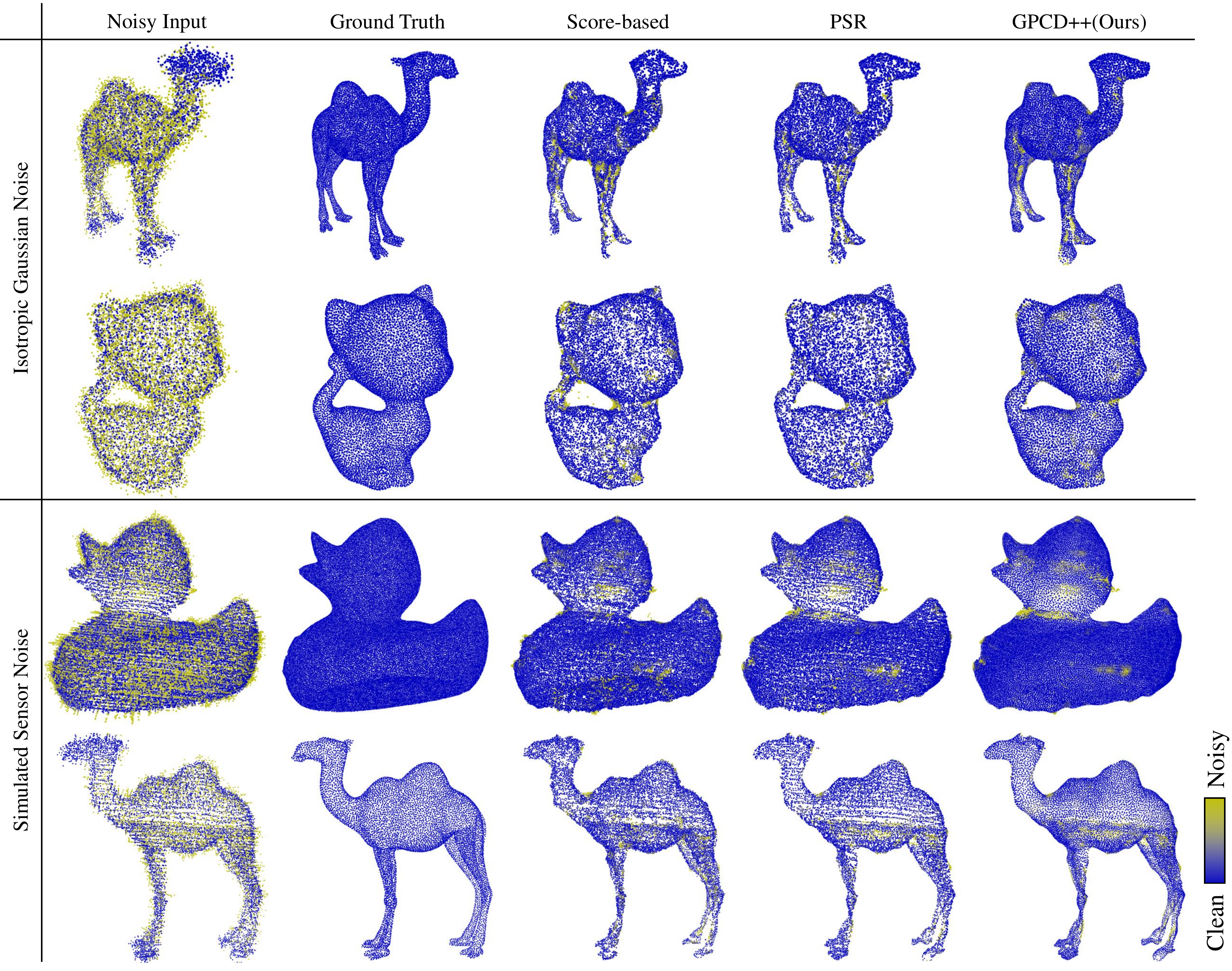}
\vspace{-0.5cm}
\caption{\textbf{Comparison of point cloud denoising methods under Isotropic Gaussian noise and simulated sensor noise.}  Our network significantly improves distribution uniformity while achieving state-of-the-art denoising performance.
}
\label{fig:visual}
\end{figure*}

We conduct qualitative studies on PU dataset and then visualize the denoising results generated by our proposed GPCD++(PSR) and previous state-of-the-art methods. Due to the limit of space, we only compare our proposed methods with Score-based Denoising Network and Point Set Resampling for the following reasons, (1) they have a significantly better performance on PU dataset than other methods, (2) they are gradient-based methods closely related to our method. Specifically, the input point cloud is perturbed by isotropic Gaussian noise with a scale parameter 3\%, and perturbed by simulated sensor noise with a scale parameter 1\%. 
As seen in Fig.~\ref{fig:visual}, although both Score-based Denoising Network and Point Set Resampling successfully converge each noisy point towards the underlying surface, the denoised point clouds suffer from distribution non-uniformity. Our method can effectively resolve this problem and produce higher-quality point clouds.

\begin{figure*}[!t]
\centering
\includegraphics[width=1.0\linewidth]{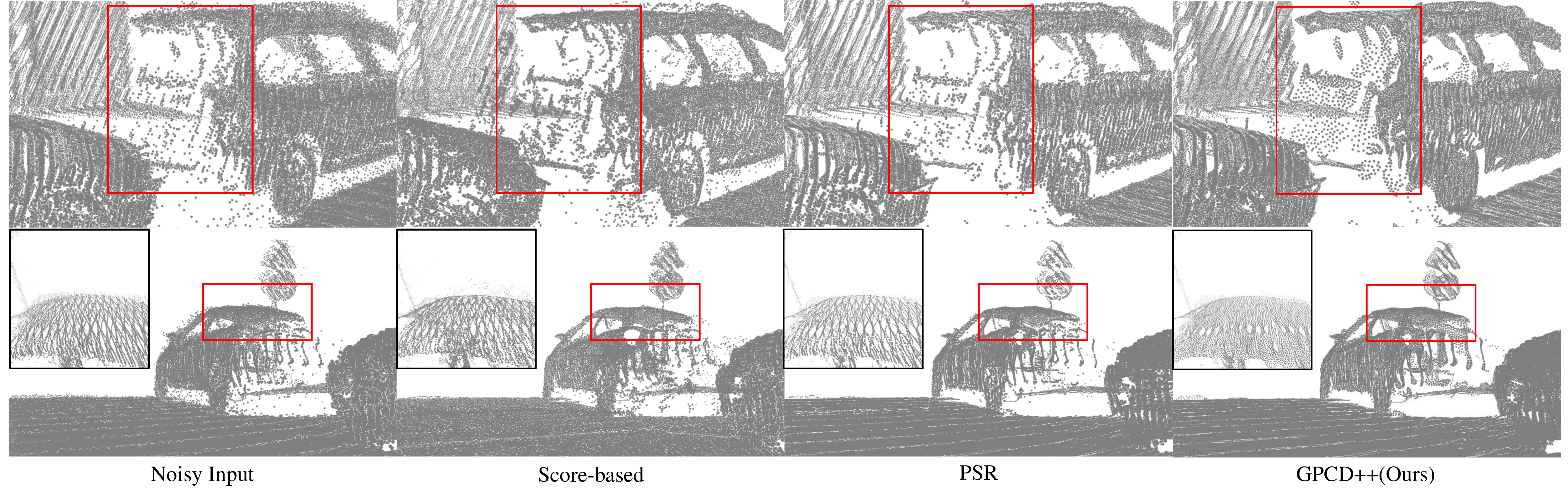}
\vspace{-0.5cm}
\caption{\textbf{Comparison of point cloud denoising methods on the real-world dataset Paris-rue-Madame~\cite{paris-rue-madame}.} Our method can preserve subtle structures and achieve distribution uniformity.}
\label{fig:visual3}

\end{figure*}

Fig.~\ref{fig:visual3} shows the denoising results of various methods on the real-world dataset Paris-rue-Madame~\cite{paris-rue-madame}. It is worth noting that our method can not only generate clean point clouds from the noisy input, but also preserve subtle structures and achieve distribution uniformity. 



\subsection{Ablation Study}




\setlength{\tabcolsep}{1.5pt}

\begin{table}
\vspace{-0.5cm}
\begin{center}
\caption{\textbf{Ablation study under isotropic Gaussian noise on PU dataset.} CD and P2M are multiplied by $10^4$.}
\label{ablation}
\begin{tabular}{c|c|cccccc}
\hline
\multicolumn{2}{c|}{\# Points} & \multicolumn{6}{c}{10K(Sparse)}\\
\multicolumn{2}{c|}{Noise} & \multicolumn{2}{c}{1\%} & \multicolumn{2}{c}{2\%} & \multicolumn{2}{c}{3\%}\\

Method & \# Parameters & CD & P2M & CD & P2M & CD & P2M \\
\hline
A:K=4,L=2 & 0.286M & 2.090 & 0.281 & 3.017 & 0.702 & 3.775 & 1.230\\ 
{Ours:K=8,L=2} & 0.286M & 1.881 & 0.251 & 2.728 & 0.654 & 3.433 & 1.161\\ 
B:K=12,L=2 & 0.286M & 1.890 & 0.256 & 2.724 & 0.661 & 3.397 & 1.154\\ 
C:K=16,L=2 & 0.286M & 1.884 & 0.254 & 2.729 & 0.659 & 3.397 & 1.150\\ 
D:K=20,L=2 & 0.286M & 1.887 & 0.255 & 2.718 & 0.659 & 3.397 & 1.152\\ 
\hline
E:K=8,L=1 & 0.275M & 1.945 & 0.289 & 2.800 & 0.697 & 3.479 & 1.168\\ 
{Ours:K=8,L=2} & 0.286M & 1.881 & 0.251 & 2.728 & 0.654 & 3.433 & 1.161\\ 
F:K=8,L=3 & 0.331M & 1.866 & 0.245 & 2.706 & 0.648 & 3.419 & 1.163\\ 
G:K=8,L=4 & 0.495M & 1.851 & 0.237 & 2.703 & 0.637 & 3.454 & 1.162\\ 
\hline
\end{tabular}
\end{center}
\end{table}

To study the impact of model size and the neighborhood size, we perform two experiments on the PU dataset. The number of nearest neighbors in KNN decides the size of local regions from which each point aggregates information. 
Tab.~\ref{ablation} shows the results of our proposed GPCD++(PSR) with different hyperparameters, where $K$ is the number of nearest neighbors in KNN and $L$ is the number of dense edge convolution layers. 
We observe no significant improvement when $K$ is larger than 8, which means UniNet only exploits local information to refine the input point clouds. The number of convolution layers decides how we reach a balance between representational power and model size. Although the larger models perform better, the improvements are moderate. Therefore, for a better trade-off between performance and model size, we choose to use $L=2$ in our implementation.

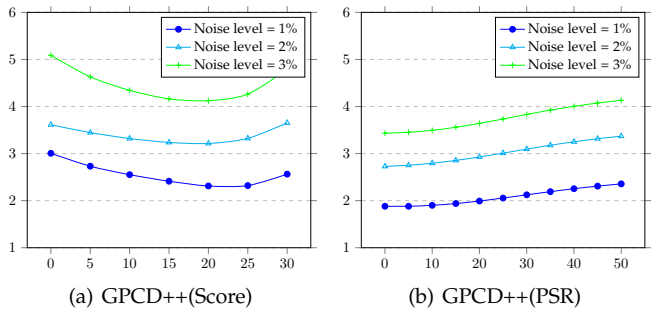
\begin{figure}[t] 
\centering 
\subfigure[GPCD++(Score)]{
\begin{tikzpicture}[scale=0.55] 
\begin{axis}[
    tick align=outside, 
    ymin=1, ymax=6,
    ymajorgrids=true,
    grid style=dashed
    ]

\addplot[smooth,mark=*,blue] plot coordinates { 
    (0,3.006)
    (5,2.732)
    (10,2.552)
    (15,2.413)
    (20,2.312)
    (25,2.320)
    (30,2.563)
};
\addlegendentry{Noise level = 1\%}


\addplot[smooth,mark=triangle,cyan] plot coordinates {
    (0,3.612)
    (5,3.446)
    (10,3.320)
    (15,3.237)
    (20,3.215)
    (25,3.323)
    (30,3.653)
};
\addlegendentry{Noise level = 2\%}


\addplot[smooth,mark=+,green] plot coordinates {
    (0,5.088)
    (5,4.630)
    (10,4.344)
    (15,4.163)
    (20,4.123)
    (25,4.262)
    (30,4.792)
};
\addlegendentry{Noise level = 3\%}

\end{axis}
\end{tikzpicture}
\label{fig:activation1}
}
\subfigure[GPCD++(PSR)]{
\begin{tikzpicture}[scale=0.55] 
\begin{axis}[
    tick align=outside, 
    ymin=1, ymax=6,
    ymajorgrids=true,
    grid style=dashed
    ]

\addplot[smooth,mark=*,blue] plot coordinates { 
    (0,1.881)
    (5,1.881)
    (10,1.901)
    (15,1.939)
    (20,1.992)
    (25,2.056)
    (30,2.124)
    (35,2.191)
    (40,2.254)
    (45,2.309)
    (50,2.357)
};
\addlegendentry{Noise level = 1\%}


\addplot[smooth,mark=triangle,cyan] plot coordinates {
    (0,2.728)
    (5,2.752)
    (10,2.795)
    (15,2.855)
    (20,2.928)
    (25,3.010)
    (30,3.095)
    (35,3.177)
    (40,3.251)
    (45,3.316)
    (50,3.370)
};
\addlegendentry{Noise level = 2\%}


\addplot[smooth,mark=+,green] plot coordinates {
    (0,3.433)
    (5,3.453)
    (10,3.494)
    (15,3.559)
    (20,3.642)
    (25,3.735)
    (30,3.832)
    (35,3.923)
    (40,4.004)
    (45,4.074)
    (50,4.132)
};
\addlegendentry{Noise level = 3\%}

\end{axis}
\end{tikzpicture}
\label{fig:activation2}
}
\caption{\textbf{Ablation study on the activation step of UniNet.} The horizontal axis represents the activation step of UniNet and the vertical axis is Chamfer Distance multiplied by $10^4$. }
\label{fig:activation}
\end{figure}

We observe noticeable performance gains from not using UniNet in the first few steps of GPCD++(Score), so we evaluate our proposed GPCD++(Score) and GPCD++(PSR) under different activation steps. As illustrated in Fig.~\ref{fig:activation1}, GPCD++(Score) achieves the best performance when the activation step is set to be 20. In contrast, Fig.~\ref{fig:activation2} shows that introducing UniNet into the gradient descent iteration of GPCD++(PSR) has a consistent improvement, and the improvement gets smaller when the activation step is larger. We assume that the difference stems from denoising efficiency of the backbone network in GPCD++. Compared with Score-based denoising net, PSR can produce high-quality denoised point clouds on early iterations, thus UniNet in GPCD++(PSR) is activated from the first step.

\section{Applications}


\begin{figure*}
\centering
\includegraphics[width=0.9\linewidth]{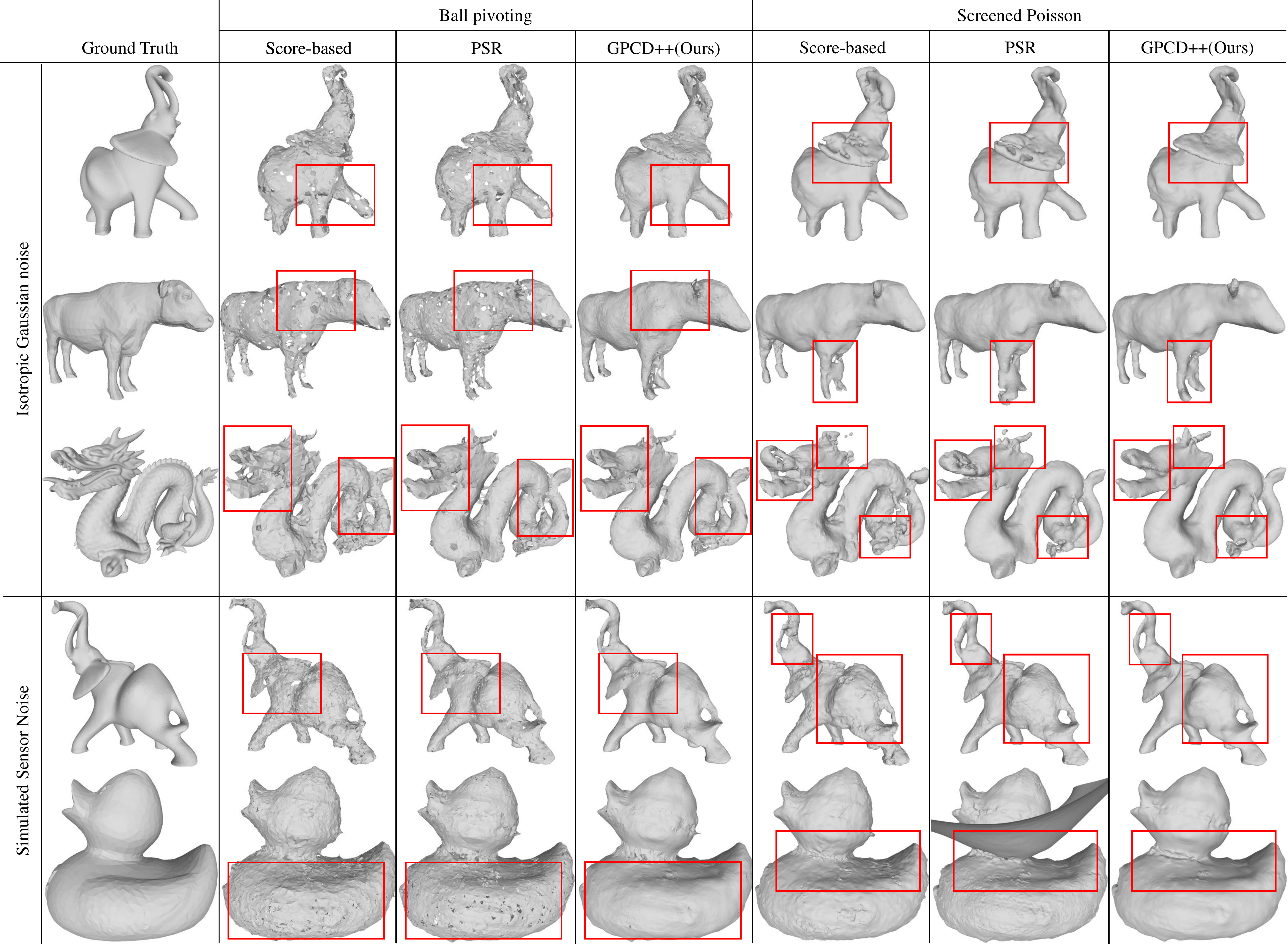}
\caption{\textbf{Comparison of surface reconstruction results with gradient-based methods.} Surface reconstruction benefits much from high-quality point clouds produced by our method.
    }
\label{reconstruction}
\end{figure*}

To demonstrate the importance of distribution uniformity, 
we further perform surface reconstruction using the denoised point clouds. Qualitative results are shown in Fig.~\ref{reconstruction}. We adopt ball pivoting~\cite{ballpivoting} and screened Poisson~\cite{screenedpoisson} for surface reconstruction. The former guarantees that the vertices of the reconstructed mesh overlap with the input point cloud, while the latter can produce a smoother mesh surface. For ball pivoting, the reconstructed surface from point clouds output by GPCD++(PSR) has much fewer holes while preserving geometric details. For screened Poisson, our results outperform others, especially on thin planes and cylinder-like parts, where normal estimation is more sensitive to distribution uniformity. Moreover, our method significantly outperforms baseline methods in fine structures and complex regions, where the reconstructed details are more sensitive to point distribution. We can also observe that our method can achieve better denoising results under simulated sensor noise and produce a smoother mesh surface. To summarize, our method consistently surpasses gradient-based methods with the two commonly-adopted surface reconstruction approaches.  

\begin{figure*}[!t]
\centering
\includegraphics[width=0.9\textwidth]{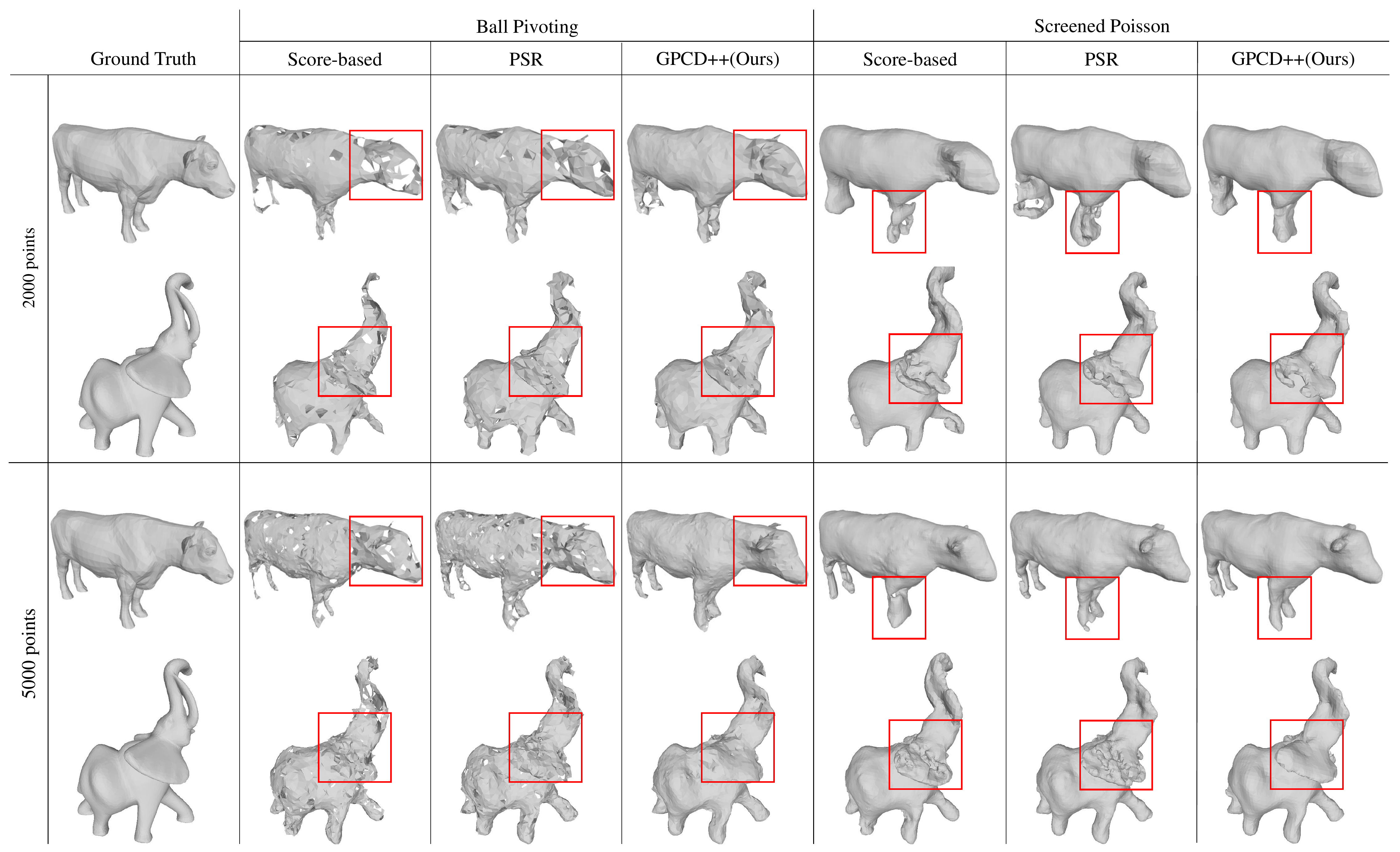}
\caption{\textbf{Comparison of surface reconstruction results with gradient-based methods under different numbers of points.}}
\label{fig:npts_rec}
\end{figure*}

The number of points determines the detailing representational power of point clouds, which further decides the quality of reconstructed meshes. Thus, we perform more qualitative experiments to explore whether our proposed GPCD++ brings consistent improvement on gradient-based denoising methods under different numbers of points. As shown in Fig.~\ref{fig:npts_rec}, the meshes reconstructed by Ball Pivoting~\cite{ballpivoting} from GPCD++(PSR) have fewer holes than that of other denoising methods. Besides, for screened Poisson surface reconstruction~\cite{screenedpoisson}, the meshes from our methods contain geometric details of higher quality. We also observe that screened Poisson method reconstructs correct surfaces of the elephant's ears using our proposed GPCD++(PSR) when the number of points is set to be 5000. In contrast, other denoising methods cannot generate high-quality point clouds to restore geometric details in the same case.  

\begin{figure}[!t]
\centering
\includegraphics[width=1.0\linewidth]{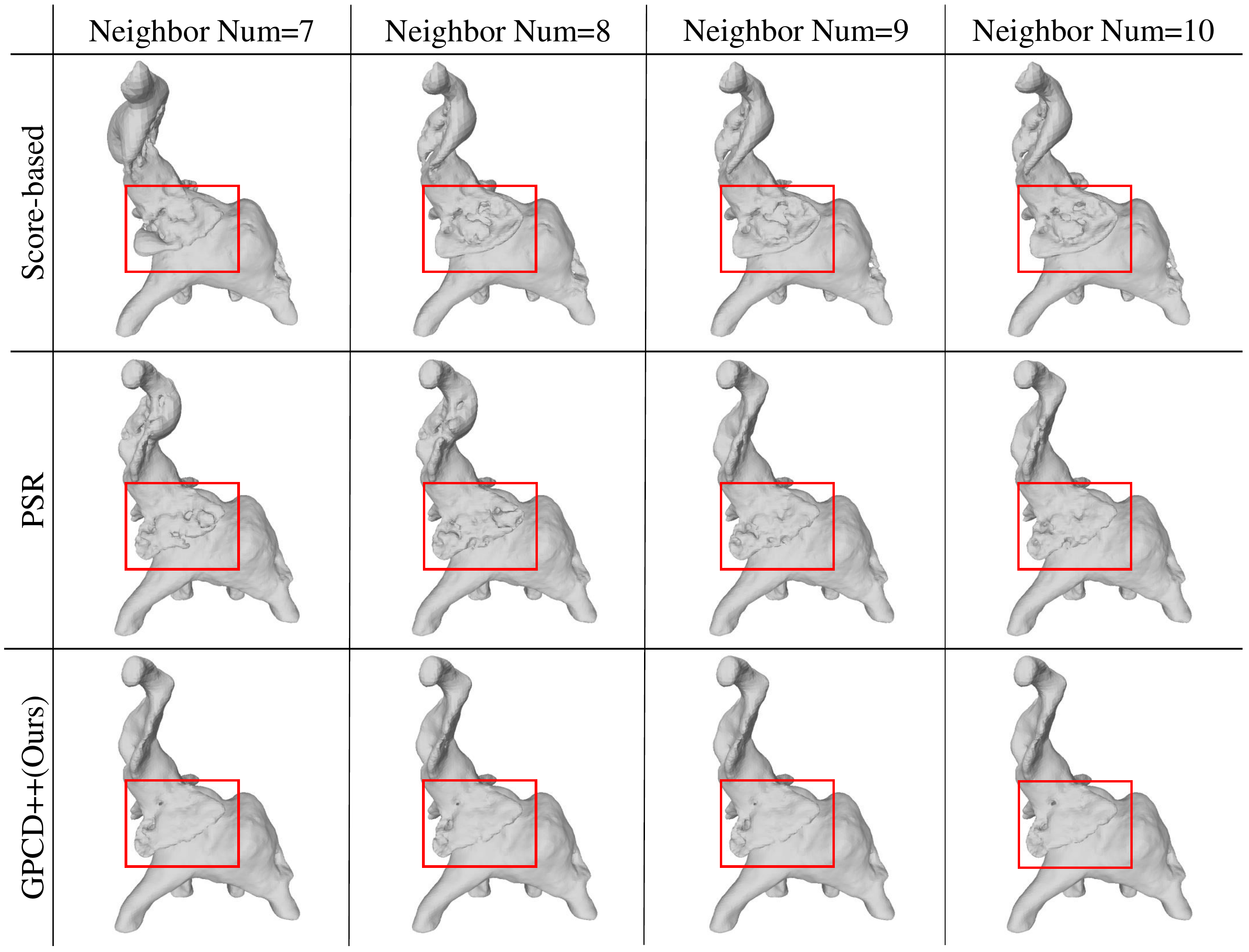}
\caption{\textbf{Comparison of screened Poisson surface reconstruction results under different hyperparameters.}}
\label{fig:nn_rec}
\end{figure}

Notably, the screened Poisson algorithm requires the normal of each vertex to create watertight surfaces from oriented point sets, thus the quality of reconstructed surface depends partly on the accuracy of normal estimation. To explore how normal estimation influences the quality of reconstructed surfaces, we choose different neighbor numbers to estimate the normal vectors and the results are shown in Fig.~\ref{fig:nn_rec}. We observe that the denoised point clouds generated by our proposed methods are more robust to hyper-parameter selection than that of other methods. Besides, the surface holes do not disappear with the change of hyper-parameters, which further demonstrates the importance of distribution uniformity.

\section{Limitations}

As shown in Fig.~\ref{method}, each point only moves in the local region of its original position, thus GPCD++ can only achieves local uniformity and fails to deal with large holes and global distribution non-uniformity, which are often caused by occlusion and limited sensor resolution. Moreover, the improvements of our method are limited for very dense point clouds compared with gradient-based denoising methods. If the noisy point cloud is very dense(\textgreater 100K points for camel in Fig.~\ref{fig:visual}), the reconstruction methods are likely to have similar results.

\section{Conclusion}

In this paper, we propose a novel point cloud denoising pipeline named GPCD++ to resolve point distribution non-uniformity in gradient-based methods. We provide a deep analysis of existing methods and demonstrate that this issue comes from the point independence assumption. Our proposed GPCD++ consists of three components, including a feature extraction unit and a gradient field estimation unit to project noisy points towards the underlying surface, along with an ultra-lightweight network named UniNet to model the interactions among points for local distribution uniformity. Quantitative experiments show that our network achieves state-of-the-art performance on benchmark datasets. Qualitative experiments demonstrate that downstream tasks such as surface reconstruction can benefit greatly from the proposed pipeline. In the future, we would like to generalize the uniformity such that it can be adaptive to surface features for further improving surface reconstruction quality. 

\ifCLASSOPTIONcompsoc
  \section*{Acknowledgments}
\else
  \section*{Acknowledgment}
\fi

The authors would like to thank Haolan Chen for his continued support and fruitful discussions. This work was supported by the Natural Science Foundation of China (Project Number 61832016), and Tsinghua-Tencent Joint Laboratory for Internet Innovation Technology.

\ifCLASSOPTIONcaptionsoff
  \newpage
\fi



%



\bibliographystyle{IEEEtran}
\bibliography{egbib}

%

\begin{IEEEbiography}[{\includegraphics[width=1in,height=1.25in,clip,keepaspectratio]{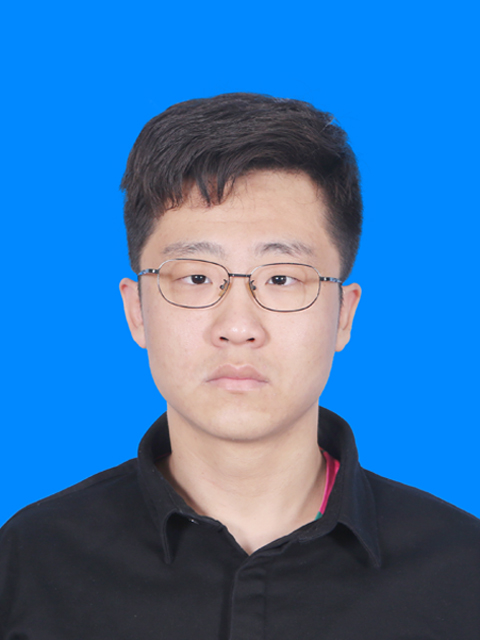}}]{Tian-Xing Xu} received his bachelor’s degree from Tsinghua University in 2021, where he is currently pursuing the PhD degree in the Department of Computer Science and Technology. His research interests include 3D computer vision.
\end{IEEEbiography}

\begin{IEEEbiography}[{\includegraphics[width=1in,height=1.25in,clip,keepaspectratio]{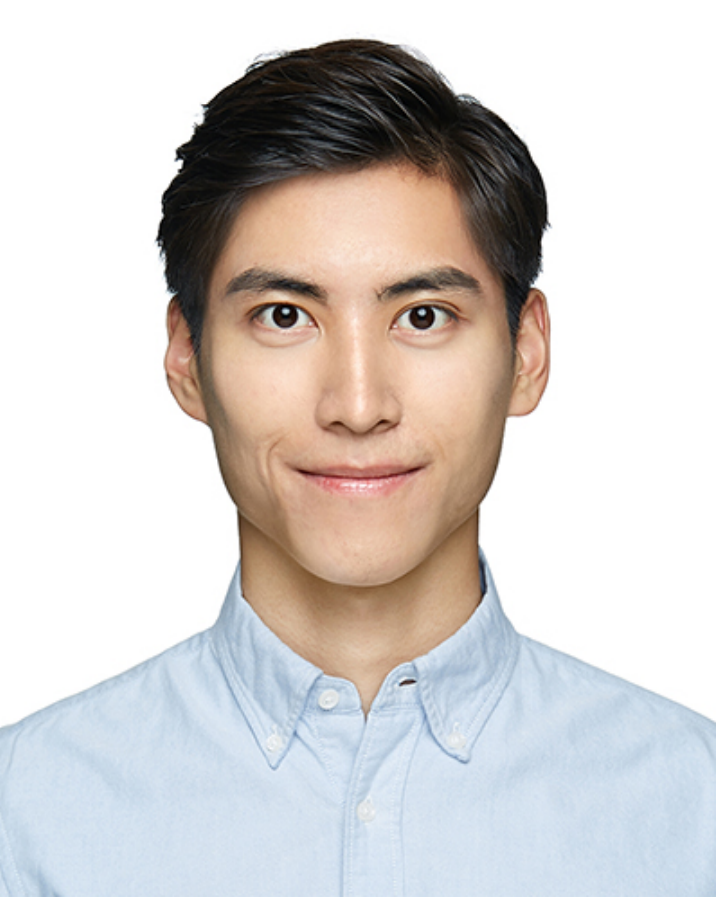}}]{Yuan-Chen Guo} received his bachelor’s degree from Tsinghua University in 2019, where he is currently pursuing the PhD degree in the Department of Computer Science and Technology. His research interests include computer graphics and computer vision.
\end{IEEEbiography}

\begin{IEEEbiography}[{\includegraphics[width=1in,height=1.25in,clip,keepaspectratio]{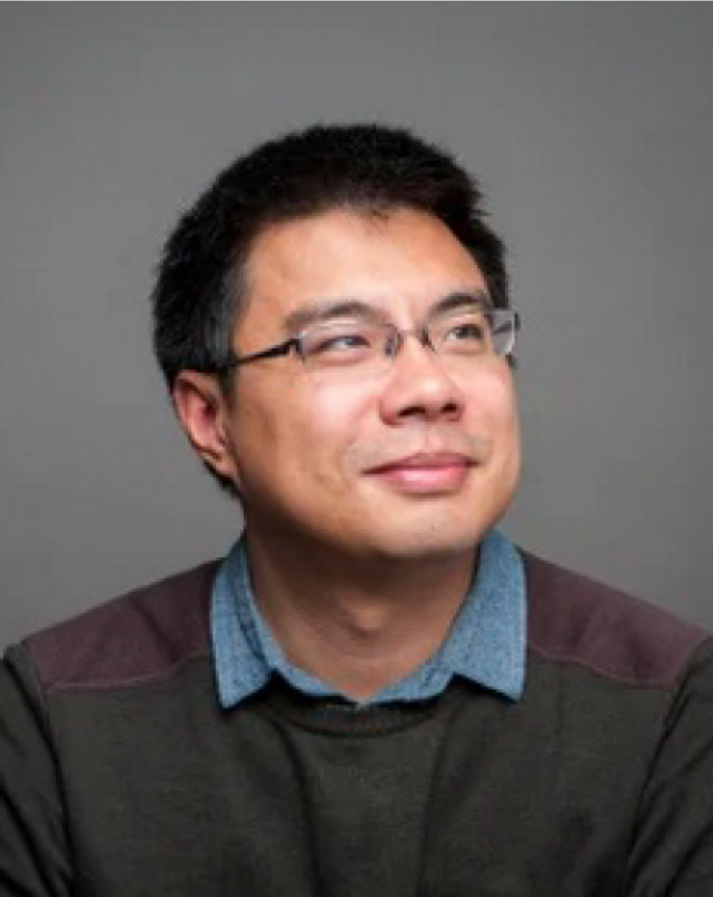}}]{Yong-Liang Yang} is a Senior Lecturer in the Department of Computer Science, University of Bath. He received the B.S. degree and the Ph.D. degree of Computer Science from Tsinghua University. His research interests are broadly in visual computing and interactive techniques.
\end{IEEEbiography}

\begin{IEEEbiography}[{\includegraphics[width=1in,height=1.25in,clip,keepaspectratio]{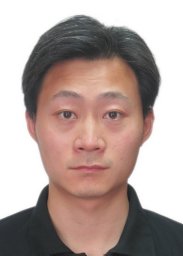}}]{Song-Hai Zhang} received the PhD degree of Computer Science and Technology from Tsinghua University, Beijing, in 2007. He is currently an associate professor in the Department of Computer Science and Technology at Tsinghua University. His research interests include computer graphics, virtual reality and image/video processing.
\end{IEEEbiography}






\end{document}